\documentclass[11pt]{article}

\usepackage[preprint]{acl}

\usepackage{times}
\usepackage{latexsym}

\usepackage[T1]{fontenc}

\usepackage[utf8]{inputenc}

\usepackage{microtype}

\usepackage{inconsolata}

\usepackage{graphicx}

\usepackage{booktabs}
\usepackage{multirow}
\usepackage{siunitx}
\usepackage{makecell}
\usepackage{subcaption}
\usepackage{float} 
\usepackage{placeins}
\usepackage{xcolor}
\usepackage{amsfonts}
\usepackage{amsmath}
\usepackage{fvextra}
\usepackage{tabularx}
\usepackage{utils}

\usepackage{enumitem}
\setlist[itemize]{noitemsep}

\newcommand{\exclusive}[1]{\textcolor{cyan}{#1}}
\newcommand{\nonexclusive}[1]{\textcolor{magenta}{#1}}

\newcommand{\qualmeta}[1]{{\footnotesize\color{gray!75!black}#1}}

\title{Not Just an Evaluation: A Case Study of Discourse Particle Sensitivity in Language Models}

\author{
 \textbf{William Sheffield\textsuperscript{1,*}},
 \textbf{Jiahui Chen\textsuperscript{1,*}},
 \textbf{Amy Zhang\textsuperscript{1}},
\\
 \textbf{Alex Warstadt\textsuperscript{2}},
 \textbf{Junyi Jessy Li\textsuperscript{1}},
 \textbf{Kanishka Misra\textsuperscript{1}}
\\
 \textsuperscript{1}The University of Texas at Austin\quad
 \textsuperscript{2}University of California, San Diego
\\
 \small{\textsuperscript{*}Equal contribution.}
\\
 \small{
   \textbf{Correspondence:} \href{mailto:sheffieldw@utexas.edu}{\texttt{sheffieldw@utexas.edu}},
   \href{mailto:jiahui.k.chen@utexas.edu}{\texttt{jiahui.k.chen@utexas.edu}}
 }
\\
 \small{
   \raisebox{-0.2\height}{\includegraphics[height=1.1em]{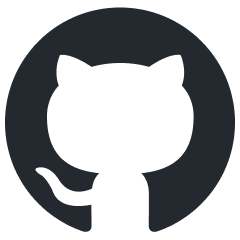}}~\href{https://github.com/JiahuiKChen/Not_Just_an_Evaluation}{\texttt{JiahuiKChen/Not\_Just\_an\_Evaluation}}
 }
}

\begin{document}

\setlength{\abovedisplayskip}{0.15em}
\setlength{\belowdisplayskip}{0.25em}
\setlength{\abovedisplayshortskip}{0.15em}
\setlength{\belowdisplayshortskip}{0.15em}

\maketitle
\begin{abstract}
Discourse particles subtly shape the flow of discourse by reflecting conversational context and interlocutor attitudes, often leaving the truth-conditional meaning untouched. 
They are also often polyfunctional, with distinct senses which partially overlap with other particles; for example, the exclusive use of English \emph{just} and \emph{only} (John only/just drinks water).
Despite the frequency of these words in natural language, they have gone largely unstudied due to the subtle nature of their effects.
We investigate LLM sensitivity to \emph{just} and \emph{only} in natural conversations using behavioral and mechanistic analyses.
First, we find that LLMs and humans similarly prefer the presence of these particles in context, and both are more sensitive to \emph{just} compared to \emph{only}. 
Second, when sampling from LLMs with and without these particles, we find LLMs consistently rate generations as more likely, given their source context, demonstrating their particle sensitivity.
In both experiments, instruction tuning increases sensitivity.
Finally, we localize particle-sensitive units within LLMs and find the overlap between \emph{just} and \emph{only} is asymmetric: \emph{only} units transfer selectively to exclusive \emph{just}, while all \emph{just} units transfer to \emph{only} equally well regardless of sense, suggesting LLM lexical-level particle representations are partially organized by sense.
We release our code, sense-labeled dialogue stimuli, and human preference annotations at \url{https://github.com/JiahuiKChen/Not_Just_an_Evaluation}.
\end{abstract}

\begin{figure*}
    \centering
    \includegraphics[width=\textwidth]{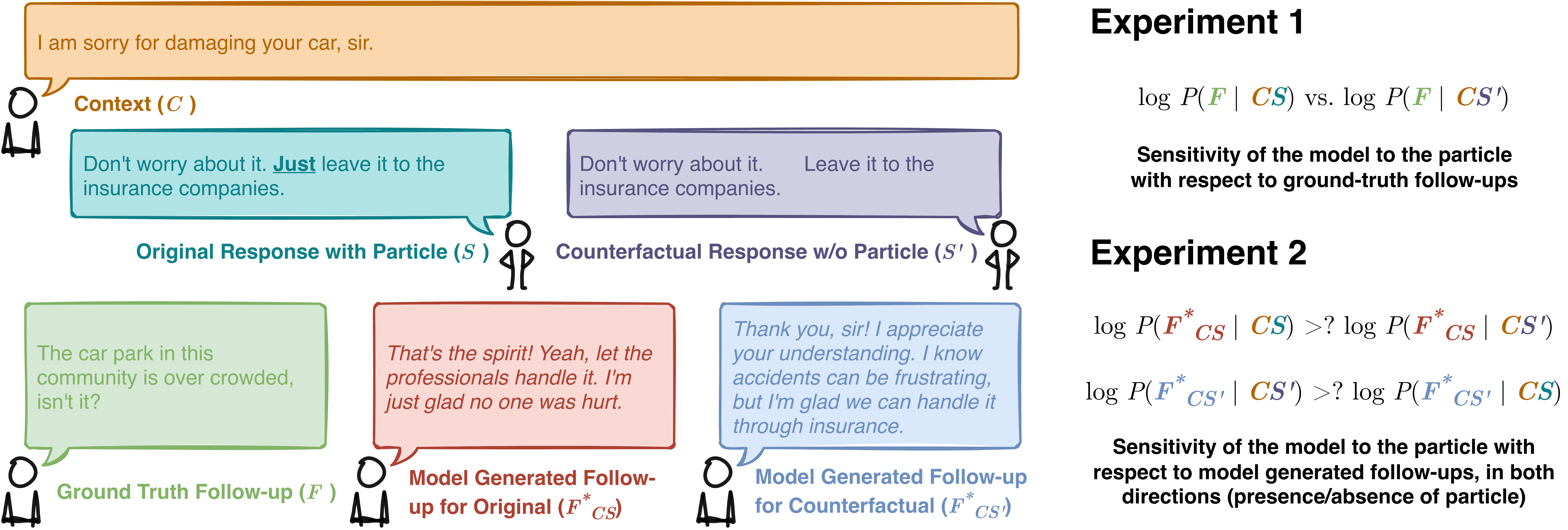}
    \caption{An example stimulus from our data, involving the context $C$, target sentence with particle $S$ and counterfactual sentence without particle $S'$, ground-truth follow-up $F$, and LLM follow-ups generated with the particle $F^*_{CS}$ and without $F^*_{CS'}$.}
    \label{fig:just_method}
\end{figure*}

\section{Introduction}

While waiting for Bob to arrive at the theater, you and Claire discuss his potential tardiness: 
\begin{itemize}
    \item[] You: Are you mad at him?
    \item[] Claire: He's (just) late all the time!
\end{itemize}

\noindent
Whether or not Claire uses \emph{just} in her response does not change Bob's habitual tardiness; rather, it subtly shapes the future discourse: without \emph{just}, you can easily question further (e.g. ``Oh does this happen often?''), while with \emph{just}, further inquiry seems to push back on Claire's claim that Bob is ``just late''. 
In this dialog, \emph{just} is an example of a \textbf{discourse particle}: a word that subtly shapes the flow of discourse and reflects conversational context and interlocutor attitudes while leaving the truth-conditional meaning largely untouched \cite{Grosz2020DiscourseP}. 
Since the acceptability of discourse particles depends on conversational norms (such as politeness, emotion, and detail) rather than felicity or grammaticality, we use \emph{sensitivity} to refer to their effects on the surrounding conversation: how well they fit the prior context, and how they shape what comes next.

Another noteworthy aspect of discourse particles is that many are polyfunctional, having multiple distinct yet interconnected meanings which often partially overlap with other particles forming constellations of meaning \cite{Aijmer+2002}.
One such ``constellation'' is the group of \emph{exclusive} particles (e.g. \emph{only}, \emph{just}, \emph{solely}, \emph{simply}): e.g., Claire saying ``I only/just\footnote{The exclusive reading is one of several available for \emph{just}} like popcorn'' excludes other salient movie snacks from her liking \cite{rooth1992focus}.
While these particles share some uses, they do not completely overlap: Claire responding ``He's only late all the time.'' in the aforementioned dialog is difficult to parse in some dialects and \emph{only} is not interchangeable with \emph{just}.
Exclusive particles have been studied extensively in Semantics \cite{horn1969presuppositional, beaver2008sense}---e.g.,\citet{Coppock2014PrinciplesOT} propose a unified account for their meaning, a common core denotation for all exclusive particles.

Given the rise of LLMs and their acquisition of a variety of linguistic phenomena as tested in minimal-pair settings \cite[][\emph{i.a.}]{linzen-etal-2016-assessing, marvin-linzen-2018-targeted, warstadt-etal-2020-blimp-benchmark, misra-etal-2023-comps}, it is natural to ask if they have also acquired sensitivity to these particles. 
This is further underscored by the frequency of discourse particles in conversation, a domain of ever-increasing importance for LLMs (especially as conversational agents). Finally, LMs are perhaps our strongest model of language \textit{use}, so testing them on polyfunctional discourse particles in terms of how they affect upcoming utterances can give us empirical evidence about their theoretically-discussed effects, bridging the gap between LLMs and Linguistic theory \cite{warstadt2024artificialneuralnetworkstell, Futrell2026-FUTHLL, misra2026systematicframeworkgeneratingnovel}.

To this end, we investigate sensitivity of LMs to selected English discourse particles \emph{just} and \emph{only} using two behavioral experiments, including a comparison to human preferences. We then probe models for evidence of a unified representation for \textit{exclusives}.
We extract occurrences of our target words and their surrounding context from the natural conversations of DailyDialog \cite{li-etal-2017-dailydialog}---see Figure~\ref{fig:just_method} for an example.
We observe that 1) LLMs, especially instruct-tuned ones, and humans similarly prefer the ground-truth conversations with the particle over counterfactual versions with the particle removed (Section~\ref{sec:particle_removal_exp}); 2) sampled follow-ups are consistently more probable under their source context (with/without the target words), showing particle sensitivity within models' distributions (Section~\ref{sec:generation}); 3) LLMs localize these particles lexically, and that some of this representation is shared between \emph{only} and \exclusive{exclusive \emph{just}} in models' residual layer, though its effects are largely asymmetric between the two.

This work builds ground for future explorations into pragmatically-sensitive continuations, which, to our knowledge, has gone understudied at the level of \textit{discourse particles}. 
Work exists for high-level pragmatic relations \cite[e.g. irony, metaphor;][]{hu-etal-2023-fine}, implicature \cite{ruiz2023implicature}, discourse connectives \cite{pandia-etal-2021-pragmatic, brubaker-etal-2026-wugnectives}, and particle sense identification \cite{sheffield-etal-2025-just} or particle prediction \cite{yusoff2026largelanguagemodelshandle}, but none of these touch on the effect of discourse particles on continuations.

\section{Background}

\paragraph{Discourse Particles}
Discourse particles have been extensively studied in Linguistics \citep{Grosz2020DiscourseP}.
The two particles we investigate, \emph{just} and \emph{only}, each have substantial lines of work.
For \emph{just}, several works analyze some of its senses and their relations \cite{lee1987semantics, beltrama2022just, wiegand2018exclusive, warstadt2020just, windhearn2021alternatives}, while \citet{Coppock2014PrinciplesOT} address \exclusive{exclusive \emph{just}} along with other exclusive particles, and \citet{thomasanddeo} propose a unified theory for 12 distinct senses.
\emph{Only} was first analyzed alongside \emph{even} by \citet{horn1969presuppositional}, and was further studied with other exclusive particles by \citet{beaver2008sense} and \citet{Coppock2014PrinciplesOT}, among others.

Most of these works either (1) unify multiple senses of a single particle into one core account or (2) unify a use shared by multiple particles into a common core; our paper draws especially from \citet{warstadt2020just}, who proposes the core effect of \emph{just} is to shape future discourse moves by communicating what the speaker views as unanswerable, and \citet{Coppock2014PrinciplesOT}, who propose a unified account for exclusive particles.

An important aspect of discourse particles is that the acceptability of their use does not always correspond to a single linguistic notion: sentences are usually grammatical and felicitous with or without them, and it usually depends on conversational norms around detail, emotion, and politeness.
For example, in responding to ``How many kids came to your niece's birthday party?'', ``Only a few.'' can give an air of frustration, ``Just a few.'' an air of sadness, while a bare ``A few.'' feels neutral or even terse.
Similarly, if a store clerk asks if they can help you, responding without the \emph{just} in ``I'm (just) browsing.'' comes off as extremely curt.
In this paper, \emph{sensitivity} refers to these differences in acceptability, which we measure through each experiment's metric.

\paragraph{Language Models \& Discourse}
The study of discourse particles using LLMs is quite new compared to the extensive work done with the related discourse connectives (e.g. \emph{however}, \emph{since}, etc.) and the discourse relations they signal. Connectives differ from discourse particle by signaling relations (e.g. ``I missed the bus, so (=RESULT) I was late this morning'') between clauses rather than speaker attitudes \cite{FRASER1999931}.
\citet{pandia-etal-2021-pragmatic} use cloze-style tests to investigate LM capabilities at predicting connectives, and \citet{brubaker-etal-2026-wugnectives} study LM capability to draw inferences from different  connectives.
Additionally, \citet{miao-kan-2025-discursive} and \citet{bhattacharyya-wein-2026-reason} use mechanistic interpretability methods to localize how LMs represent discourse relations signaled by connectives.
Both works frame their task as a prompted choice between two candidate answers, scored by the logit difference between them; in contrast, we use naturally occurring multi-turn dialogues and measure how removing a particle shifts the probability of entire follow-up turns, including the model's own generations.

Outside of discourse connectives and relations, \citet{schuster-linzen-2022-sentence} find LMs hallucinate new entities in the discourse by failing to resolve indefinite NPs in contexts such as negation; \citet{kim-misra-2026-hey} find models follow pragmatic structure by preferring continuations that are at-issue and are sensitive to cues that humans use to modulate at-issueness; \citet{hu-etal-2023-fine} study LM performance on identifying high-level pragmatic relations such as irony and metaphor, and what linguistic cues LMs use to do so; and \citet{ruiz2023implicature} study LM reasoning on implicatures that answer binary questions.
Prior work that addresses discourse particles includes \citet{sheffield-etal-2025-just}, who focus on LM capability to distinguish different senses of discourse particles in a metalinguistic setting, and \citet{yusoff2026largelanguagemodelshandle}, who prompt LMs to predict discourse particles in Colloquial Malay.

We draw from the minimal pair paradigm of \citet{linzen-etal-2016-assessing} and BLiMP \cite{warstadt-etal-2020-blimp-benchmark} and extend beyond the sentence to include pre- and post-context, following the theory of \citet{warstadt2020just}, for our first experiment (Section \ref{sec:particle_removal_exp}).
Further, our second experiment (Section \ref{sec:generation}) takes inspiration from the paradigm of \citet{kim-misra-2026-hey} where we compare generated continuations based on minimally differing contexts and applies it to study particle sensitivity rather than at-issue sensitivity.

\paragraph{Function Localization}
For our third experiment (Section \ref{sec:localization}), we adapt methods from activation patching that localize model components that show the greatest sensitivity to these particles.
Prior work has used variations of these techniques to: localize mechanisms for subject--verb agreement \cite{finlayson-etal-2021-causal}; measure recovery of the factual prediction \cite{meng-etal-2022-locating}; localize regions that identify grammatical and ungrammatical features \cite{kryvosheieva-etal-2026-different}. 
Our work extends beyond grammatical features to the pragmatics of discourse particles, and differs in methodology from \citet{kryvosheieva-etal-2026-different} by instead localizing whole attention, MLP, and residual-stream activations at particular layers and evaluating them with counterfactual activation replacement.

\section{Experimental Setup}\label{sec:exp_setup}

\subsection{Dataset}\label{sec:dataset_exp_setup}

For all our experiments, we use a subset of conversations from DailyDialog~\cite{li-etal-2017-dailydialog},\footnote{DailyDialog is released with a CC0 1.0 Universal License and is free to use with attribution to \citet{li-etal-2017-dailydialog}.} a set of natural dialogues, that contain our target words. 
We extract samples consisting of conversation context $C$, original utterance containing a target word $S$, a minimally edited utterance $S'$ with the target word removed, and the ground-truth follow-up $F$ (Figure~\ref{fig:just_method}). 
To help ensure the target word is relevant to the follow up in the conversation, we retain only examples where the target word appears in the final sentence of $S$ and the target utterance is not a question. 
We additionally restrict the dataset to conversations whose context $C$ contains at most three preceding turns, as preliminary experiments with longer contexts yielded noisier estimates of the subtle effects of word removal.

We filter for conversations where the utterance $S$ contains discourse particles \emph{just} or \emph{only}. Additionally, since discourse particle effects are subtle and do not always alter the truth-conditional semantics of the sentence, we use \emph{not} as a control since it has a clear impact on the truth-conditions.
After filtering our dataset contains $1{,}182$ \emph{just} examples, $360$ \emph{only} examples, and $1{,}307$ \emph{not} examples. 

From these data, we manually select 60 for each target word to form a gold dataset which is particularly sensitive to the presence/absence of the word.\footnote{As judged by one of the authors, a native English speaker and linguist with background in semantics and pragmatics.} 
This provides a dataset which is both more manageable for some experiments and is predicted to provide the best signal.
The \emph{just} data are split by sense into two equal sense subsets, \exclusive{exclusive} (``My picky brother just/only eats french fries and pizza'')  and \nonexclusive{non-exclusive}, for use in Section \ref{sec:localization}.

\subsection{Scoring Setup}\label{sec:scoring_exp_setup}
For probability scoring in our experiments, we convert raw \texttt{\_\_eou\_\_} markers in DailyDialog conversations into alternating \texttt{Speaker 1:}, \texttt{Speaker 2:} turn labels.
We score base models directly on this speaker-labeled text and instruction-tuned models through their native chat templates using the same dialogue content.
Following \citet{hu-levy-2023-prompting}, we measure probabilities directly rather than prompting for metalinguistic judgments, as prompting can under-estimate model knowledge.
To avoid spurious penalties from sentence-initial lower casing after deletion, we capitalize the first alphabetic character of both utterance variants before scoring.
In App.~\ref{app:prompt_examples}, we provide an end-to-end example of the $C$, $S$, $S'$, and $F$ construction, both prompt renderings, and sampled follow-ups.
We use experiment-specific scoring metrics, which we describe in the corresponding sections.

\subsection{Models}\label{sec:models_exp_setup}

We study base and instruction-tuned checkpoints from four model families: Llama~3~8B~\cite{llama3herdmodels}, OLMo-2~7B~\cite{olmo20242olmo2furious}, Qwen3.5~9B~\cite{qwen3.5}, and Gemma-2~9B~\cite{gemma2improvingopen}.
This model set keeps scale roughly comparable while spanning independently trained open-weight LLM families.
Including both base and instruction-tuned checkpoints for each model family lets us compare discourse-particle sensitivity before and after post-training.
App.~\ref{app:model_differences} summarizes documented differences in architecture, training data, and post-training across these families.



\section{Experiment 1: Particle Removal Effects for Humans and LMs}\label{sec:particle_removal_exp}


\begin{figure}
    \centering
    \includegraphics[width=\linewidth]{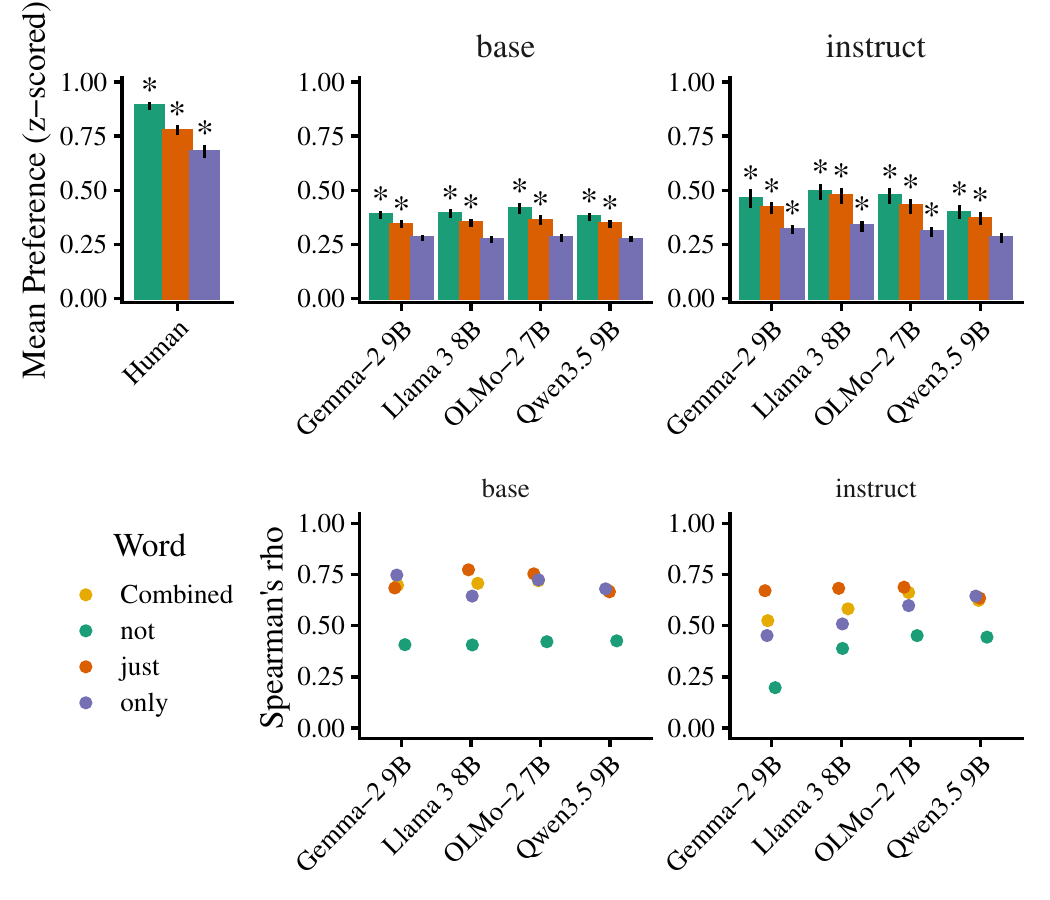}
    \caption{\textbf{Top:} Z-scored LLM and Human preferences on the gold data (distractors excluded). LLM preference is defined in Eq.~(\ref{eq:delta_log_prob_token_removal}), Human preference is rated on a 7-pt Likert scale, with 7 being strongly prefer the original passage with the word ($CSF$). Stars indicate significance on a Wilcoxon signed-rank test with Bonferroni correction.
    \textbf{Bottom:} Spearman's $\rho$ between human and LLM ratings, reported per word as well as for all words.}
    \label{fig:particle_removal}
\end{figure}


We begin investigating LLM sensitivity to discourse particles by comparing LM preferences to those of human participants. We measure LM and human preference between $CSF$ and $CS'F$ on the gold dataset (Section~\ref{sec:dataset_exp_setup}), examining \emph{just} and \emph{only} with \emph{not} as a meaning-altering control. 
Models that effectively encode these words’ conversational contribution should assign higher probability to the original dialogue than to the edited variant.

\subsection{Scoring Metrics}
\paragraph{LMs}
For models, our conversation-level score is the log-probability analogue of the comparison introduced above:
\begin{equation} \label{eq:delta_log_prob_token_removal}
    \Delta_{\mathrm{removal}} = \log P(CSF) - \log P(CS'F)
\end{equation}
Positive values indicate that the model prefers the ground-truth conversation to the counterfactual one.
Since $S$ and $S'$ have different token lengths, all reported deltas are normalized per token. 

\paragraph{Humans}
In order to make the preference task for humans as similar to the task of the LMs, annotators are presented with $CSF$ and $CS'F$ as continuous passages in two columns (randomized) and asked to rate their preference using an implicit 7-point Likert scale slider, which closely parallels the score computed for LMs. 
To establish baseline annotation validity, we included a set of distractor stimuli using randomly-sampled conversations from the dataset, not necessarily with one of our target words, where $S'$ was formed by randomly reordering the words in $S$. 
The distractors were randomly mixed in with the actual data.
In total, 9 participants provided preferences for the gold data (240 total, 60 x 3 target words + 60 distractors), over two equal rounds of annotation.
See App.~\ref{app:human_ann} for the annotation interface and further details.

\subsection{Results}

All human participants consistently prefer the control distractor data (median rating 7), and high agreement across the dataset.\footnote{Intraclass Correlation Coefficient with two-way random effects, absolute agreement, and averaged measures \cite{mcgraw1996icc} $ICC(A,k) = 0.854$, indicating high agreement \cite{koo2016iccinterp}.}
The top of Figure~\ref{fig:particle_removal} shows LM and human preference ratings on the gold data. 
Humans show a preference for all the target words, and models follow this trend for \emph{not} and \emph{just}.
The preferences for ``only'' are less strong for humans, and for all base models and one instruct model, the preference is not significantly above zero.
We also observe generally strong positive correlations of models with humans across all words, and especially with \emph{only} and \emph{just}. 
The lower correlation with \emph{not} appears to fall out of the concentrated distribution of human labels (they show strong, categorical-like preference) contrasted with the more graded LM ratings. 
Ultimately, we find humans and models qualitatively agree in their continuation preferences with respect to particles, with post-training increasing this sensitivity.



\section{Experiment 2: Generation}
\label{sec:generation}
\label{sec:llm_generation}

\begin{figure}[t]
\centering
\includegraphics[width=\linewidth]{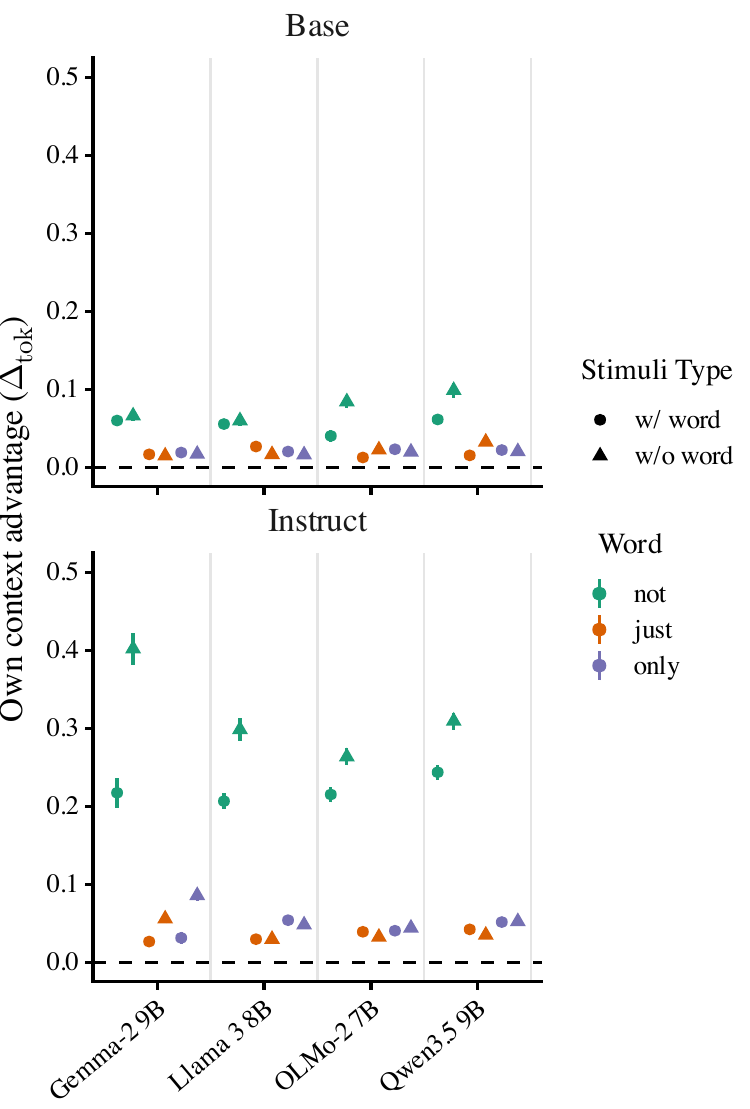}
\caption{Own-context advantage by model in the generation experiment. We construct paired prompt contexts $S$ and $S'$ that differ only in the target word, sample next-turn responses from both prompts, and then rescore each sampled response under both contexts. $\Delta_{\mathrm{tok}}$ is the mean per-token difference (Equation~\ref{eq:delta_token})
Positive values indicate an own-context advantage.}
\vspace{-0.2cm}
\label{fig:generation-own-context-advantage-by-model}
\end{figure}

Section~\ref{sec:particle_removal_exp} explored whether models penalize a minimally edited utterance when a target word is removed and compare to a ground-truth follow-up. This leaves two gaps: First, it paints an asymmetric picture of model sensitivity---we can only measure model preference for follow-ups that are appropriate only \textit{with} the particle/word present. Second, we restrict the model to a single continuation, when models likely represent their own distribution of follow-up continuations \citep{kim-misra-2026-hey}.
To address these, we turn to analyses that incorporate models own generated input, both in the presence and absence of a given particle/word.
For each item, we compare paired prompt contexts $CS$ and $CS'$ that differ only in the presence of the target word, sample continuations from both versions, and then test whether each continuation is more probable under the context it came from than under the alternate context. 
If a model represents the conversational contribution of a discourse particle, responses sampled from the particle-bearing prompt $CS$ should show an own-context advantage relative to the particle-removed prompt $CS'$, and vice versa. 
Put another way, models that are sensitive to these particles adjust their follow-up generations based on the presence and absence of the particle.
We run this procedure for \textit{just} and \textit{only}, with \textit{not} as a meaning-altering control.

\subsection{Methodology}
For each stimulus item (see Section~\ref{sec:dataset_exp_setup}), the sampling prompts are the observed conversation $CS$ and its counterfactual $CS'$, where the target word is deleted from a naturally occurring utterance. 
For base models, prompts are rendered as speaker-labeled conversations ending at the next-speaker prefix. 
For instruction-tuned models, we use each tokenizer's native chat template, placing the target utterance as the final user turn so that the model generates the next assistant response, without adding a task-specific system prompt (see App.~\ref{app:prompt_examples} for a complete example). 

From each of $CS$ and $CS'$, we sample $50$ unique next-turn responses using nucleus sampling \cite{holtzman2020curiouscaseneuraltext} with top-p 0.9 and temperature of 1.0, truncating generations at the first speaker-turn boundary.\footnote{Prior work suggests that 50 samples offer a strong compute--performance tradeoff \cite{giulianelli-etal-2023-information}.}
Each sampled response $y$ is then re-scored under both prompt variants. 
Let $CS_{\mathrm{src}}$ denote the prompt from which $y$ was sampled and $CS_{\mathrm{alt}}$ the matched alternate prompt. We compute an own-context advantage as:

\begin{equation}
\label{eq:delta_token}
\vspace{1em}
\Delta_{\mathrm{tok}}(y)=\bar{\ell}(y \mid CS_{\mathrm{src}})-\bar{\ell}(y \mid CS_{\mathrm{alt}}),
\end{equation}

\noindent
where $\bar{\ell}(y \mid CS)$ is the sequence log-probability of $y$ under prompt $CS$, normalized by the number of generated tokens. 
If $y$ is sampled from $CS$, we compute $\bar{\ell}(y \mid CS) - \bar{\ell}(y \mid CS')$; if $y$ is sampled from $CS'$, we reverse the subtraction so that positive $\Delta_{\mathrm{tok}}$ values always favor the source context.

\subsection{Results}
Figure~\ref{fig:generation-own-context-advantage-by-model} shows that changing the target word changes the continuation space the model expects: in all model--word--source combinations, sampled responses are more probable under the prompt they were generated from than under the matched alternate prompt. As in Experiment~1, the largest effects come from \textit{not}.
Effects for discourse particles are roughly an order of magnitude smaller (Table~\ref{tab:candidate_ci}) and, we see generally consistent preferences regardless of the edit. Overall, models do encode discourse particles in next-turn expectations, but more subtly compared to a truth-conditional edit like removing \textit{not}.

\paragraph{Qualitative examples}
Table~\ref{tab:qual_examples} shows follow-ups sampled from instruction-tuned LLMs.
In (a), \emph{just} conveys that the customer does not want assistance: the follow-up sampled from $CS$ withdraws the offer of help, as the gold follow-up $F$ does, whereas the follow-up sampled from $CS'$ requests further details about the bag.
In (b), the follow-up sampled from $CS$ reflects the implication of \emph{only} that other laundry remains, as does $F$, whereas the follow-up sampled from $CS'$ treats the laundry as finished.
Not every item elicits distinct follow-ups, however.
In (c), Llama~3~8B Instruct responds with nearly identical praise, with or without \emph{just}, and its own-context advantages are correspondingly small.
In both cases, it responds in the register of a helpful assistant rather than as the shopkeeper who produces $F$, so its follow-ups can differ from human ones for reasons unrelated to the particle.
App.~\ref{app:qual_examples} provides examples from other LLMs.

\begin{table*}[]
\centering
\footnotesize
\renewcommand{\arraystretch}{1.15}
\begin{tabularx}{\textwidth}{@{}c>{\raggedright\arraybackslash}X>{\raggedright\arraybackslash}p{0.15\textwidth}>{\raggedright\arraybackslash}X>{\raggedright\arraybackslash}X@{}}
\toprule
 & Context $C$ and utterance $S$ / $S'$ & Gold follow-up $F$ & Sampled from $CS$ ($F^{*}_{CS}$) & Sampled from $CS'$ ($F^{*}_{CS'}$) \\
\midrule
(a) & \textbf{C:} \ldots I'm looking for a black, leather bag. / How's this? \newline \textbf{S:} I'm (just) looking. \newline \qualmeta{Human: \textbf{6.2}} & If you need any help, let me know. & Understood! No pressure. & I understand! I need a little more information to help you find the perfect black leather bag. \\
 & & & \multicolumn{2}{l}{\qualmeta{\textit{Gemma-2 9B IT}; $\Delta_{\mathrm{tok}}$: $0.75$ ($CS$), $-0.03$ ($CS'$)}} \\
\midrule
(b) & \textbf{C:} Did you already wash your clothes today? \newline \textbf{S:} I've (only) washed my white clothes. \newline \qualmeta{Human: \textbf{6.4}} & Could you wash some of my dark clothes with yours? & Oh, okay! Are you planning on washing the rest of your clothes later? & Good job! \\
 & & & \multicolumn{2}{l}{\qualmeta{\textit{Gemma-2 9B IT}; $\Delta_{\mathrm{tok}}$: $0.17$ ($CS$), $0.51$ ($CS'$)}} \\
\midrule
(c) & \textbf{C:} Hello! Are you shopping here today? Haven't I seen you in some flea markets and thrift shops a couple of times? \newline \textbf{S:} I'm (just) comparing prices. \newline \qualmeta{Human: \textbf{5.9}} & You should! This thrift shop is really cheap, though, you have to pay cash here. & A savvy shopper! Comparing prices is a great way to make sure you're getting the best deal. Are you looking for something in particular or just browsing? & A savvy shopper! Comparing prices is a great way to make sure you're getting the best deal. Are you looking for something in particular or just browsing to see what catches your eye? \\
 & & & \multicolumn{2}{l}{\qualmeta{\textit{Llama~3~8B Instruct}; $\Delta_{\mathrm{tok}}$: $0.05$ ($CS$), $0.02$ ($CS'$)}} \\
\bottomrule
\end{tabularx}
\caption{Follow-ups sampled from instruction-tuned LLMs for three gold items. The target word appears in parentheses in $S$ and is removed in $S'$; ``/'' separates turns in $C$. Human is the mean rating on the 7-point scale (7: strongly prefer $CSF$). Each sampled follow-up is selected for illustration from the 50 samples per history; $\Delta_{\mathrm{tok}}$ is the LLM's mean own-context advantage (Equation~\ref{eq:delta_token}) over all 50 samples from each history.}
\vspace{-0.4cm}
\label{tab:qual_examples}
\end{table*}

\section{Experiment 3: Function Localization Analysis}
\label{sec:localization}

The previous experiment showed that removing a discourse particle changes the continuations a model expects, but this behavioral effect does not reveal which internal computations support that sensitivity, or whether models represent shared structure across particles/words. 
Experiment~3 therefore asks whether the effect can be localized to particular layer--component sites and whether those sites are causally involved in preserving particle-sensitive response probabilities. 
We first localize sites for \textit{not}, \textit{just}, and \textit{only} (Section~\ref{sec:exp3a_particle_loc}), then test whether localization transfers specifically between the \exclusive{exclusive sense of \textit{just}} and the semantically similar particle \exclusive{\textit{only}} (Section~\ref{sec:exp_3b_sense_loc}). 

\subsection{Experiment 3a: Component-Based Localization of Word Sensitivity}\label{sec:exp3a_particle_loc}

\paragraph{Setup} Our localization approach most directly adapts the \emph{causal tracing} variant of \citet{meng-etal-2022-locating}, using $CS$ and $CS'$ as minimal pairs.
We apply counterfactual activation patching to the paired DailyDialog prefixes defined in Section~\ref{sec:dataset_exp_setup}. For each item, the clean condition is the observed prefix $CS$, comprising context $C$ followed by the original utterance $S$ containing the target word; the corrupted condition is the matched prefix $CS'$, in which $S$ is replaced by its target-word-removed counterpart $S'$. 
In activation patching, an internal state produced by one input is substituted into a forward pass on another input, an intervention previously formulated as causal mediation and interchange intervention~\citep{vig-etal-2020-investigating,geiger-etal-2020-neural,heimersheim-nanda-2024-patching}. 
Since Llama 3 8B Instruct showed the highest average sensitivity in our generation experiments, we localize only on this model. 
At each of Llama~3's 32 decoder layers, we treat each of the self-attention output, MLP output, and post-block residual stream as a separate candidate localization site.



\paragraph{Localization} 
For each word, we begin with the responses generated in Section~\ref{sec:generation} from the $CS$ prefixes of the 60 manually selected DailyDialog items described in Section~\ref{sec:dataset_exp_setup}; we retain 500 responses with the largest positive own-context advantage (Equation~\ref{eq:delta_token}).
These high-advantage responses form a pool for which the model’s sensitivity to the target word is especially pronounced.
We divide each pool into two folds by DailyDialog item, keeping all responses generated from the same $CS$ prefix in the same fold. 
For each cross-validation split, localization is performed on one fold and causal evaluation on the other.
To prevent items with many selected generations from dominating localization, we use only the highest-advantage response from each item when estimating site scores.
This yields localization samples of 26 \textit{not}, 39 \textit{just}, and 32 \textit{only} items, one response per item.

For every response $y$ in a localization fold and every candidate site $s$, we first run the model on $CS$ and cache the full activation vector at site $s$ for the final token of $S$, immediately before the chat template's end-of-turn marker.
We then run the matched $CS'$ prefix and replace the activation at the same site for the final token of $S'$ with the cached $CS$ activation. Let $\bar{\ell}_{\mathrm{patch},s}(y\mid CS')$ denote the per-token log-probability of $y$ in this patched $CS'$ run. We define the site's restoration score as:

\begin{equation*}
\vspace{1em}
    R_s =
    \frac{\bar{\ell}_{\mathrm{patch},s}(y\mid CS')
    - \bar{\ell}(y\mid CS')}
    {\max\!\left(\bar{\ell}(y\mid CS)
    - \bar{\ell}(y\mid CS'), 10^{-6}\right)}.
\end{equation*}
Within each localization fold, we average $R_s$ across DailyDialog items. We then rank the 32 layer-specific sites separately within each component type by mean restoration and retain the top $k\in\{1,3,5,10\}$ sites.

\paragraph{Causal evaluation}
We evaluate the selected sites on the held-out fold using a counterfactual intervention to test for its necessity \citep{geiger-etal-2021-causal, heimersheim-nanda-2024-patching}. 
Starting from $CS$, we replace each selected site's activation at the final token of $S$ with the corresponding activation at the final token of $S'$ from $CS'$ and measure:

\begin{align*}
    \Delta_{\mathrm{patch}}
    &= \bar{\ell}\!\left(y\mid CS;
      a_s\leftarrow a_s(CS')\ 
      \forall s\in\mathcal{S}_k\right)\\
      &\hspace{3em}- \bar{\ell}(y\mid CS).
\vspace{1em}
\end{align*}

\noindent
A negative $\Delta_{\mathrm{patch}}$ value indicates that replacing the representation of the observed utterance $S$ with that of its target-word-removed counterpart $S'$ lowers the response probability, providing causal evidence that the selected sites are necessary for the model's word sensitivity. 
For each component and $k$, we compare the localized sites with four equally sized sets of randomly selected layers drawn from the remaining sites. 
Although localization uses only the highest-advantage response from each training-fold DailyDialog item, causal evaluation uses all retained responses from items in the held-out fold. We therefore first average $\Delta_{\mathrm{patch}}$ across responses from the same DailyDialog item and then across items within each fold. Table~\ref{tab:localization-merged} reports the mean of the two fold-level effects.

\paragraph{Results}
As shown in Table~\ref{tab:localization-merged}, patching the localized sites has a significantly larger effect than patching random sites in 35 of the 36 word--component--$k$ comparisons. 
The effect increases as more sites are patched, indicating that word sensitivity is both concentrated at highly informative sites and distributed across multiple layers. Residual-stream and attention sites produce the largest effects, while MLP effects are smaller. Across component types, the effects are strongest for \textit{not}, followed by \textit{only} and then \textit{just}, matching the own-context ordering for Llama~3~8B Instruct in Experiment~2. These held-out interventions therefore provide causal evidence that the localized sites support the model's word-sensitive response probabilities.
\begin{table*}[t]
\centering
\footnotesize
\setlength{\tabcolsep}{4.5pt}
\begin{tabular}[t]{cr rrrrrr @{\hspace{1.8em}} rr}
\toprule
 & & \multicolumn{6}{c}{\textbf{Localization:} $\Delta_{\text{patch}}$ ($\downarrow$ = stronger)}
 & \multicolumn{2}{c}{\textbf{Transfer:} $\Delta_{\text{sense}}$ ($+ = J_E$ stronger)} \\
\cmidrule(l{3pt}r{3pt}){3-8} \cmidrule(l{3pt}r{3pt}){9-10}
\multirow{2}{*}{Component} & \multirow{2}{*}{top-$k$}
 & \multicolumn{2}{c}{\textit{not}} & \multicolumn{2}{c}{\textit{just}} & \multicolumn{2}{c}{\textit{only}}
 & \multirow{2}{*}{\textit{just} $\rightarrow$ \textit{only}}
 & \multirow{2}{*}{\textit{only} $\rightarrow$ \textit{just}} \\
\cmidrule(l{3pt}r{3pt}){3-4} \cmidrule(l{3pt}r{3pt}){5-6} \cmidrule(l{3pt}r{3pt}){7-8}
 & & localized & random & localized & random & localized & random & & \\
\midrule
 & 1 & \textbf{-0.47} & -0.06 & \textbf{-0.22} & -0.01 & \textbf{-0.34} & -0.12 & \textbf{-0.36} & 0.01\\
 & 3 & \textbf{-2.15} & -0.41 & \textbf{-1.09} & -0.07 & \textbf{-2.10} & -0.13 & \textbf{-0.78} & 0.18\\
 & 5 & \textbf{-6.59} & -0.35 & \textbf{-1.72} & -0.12 & \textbf{-3.29} & -0.12 & -0.60 & 0.52\\
\multirow{-4}{*}{\centering\arraybackslash Attn.} & 10 & \textbf{-15.88} & -0.34 & \textbf{-2.98} & -0.10 & \textbf{-5.51} & -0.18 & 0.29 & \textbf{1.15}\\
\cmidrule{1-10}
 & 1 & \textbf{-0.46} & -0.04 & -0.26 & -0.09 & \textbf{-0.38} & -0.08 & 0.10 & 0.17\\
 & 3 & \textbf{-0.84} & -0.41 & \textbf{-0.75} & -0.08 & \textbf{-0.99} & -0.10 & 0.04 & 0.25\\
 & 5 & \textbf{-1.64} & -0.66 & \textbf{-0.87} & -0.16 & \textbf{-1.10} & -0.19 & -0.42 & 0.27\\
\multirow{-4}{*}{\centering\arraybackslash MLP} & 10 & \textbf{-3.46} & -0.24 & \textbf{-1.11} & -0.28 & \textbf{-1.34} & -0.24 & -0.24 & 0.52\\
\cmidrule{1-10}
 & 1 & \textbf{-5.31} & -0.29 & \textbf{-1.79} & -0.49 & \textbf{-3.25} & -0.84 & 0.25 & \textbf{0.84}\\
 & 3 & \textbf{-7.94} & -5.35 & \textbf{-2.23} & -0.36 & \textbf{-3.89} & -1.87 & 0.02 & \textbf{1.14}\\
 & 5 & \textbf{-12.13} & -3.67 & \textbf{-2.66} & -0.78 & \textbf{-5.15} & -1.37 & \textbf{-0.65} & \textbf{1.34}\\
\multirow{-4}{*}{\centering\arraybackslash Resid.} & 10 & \textbf{-18.32} & -1.53 & \textbf{-3.45} & -0.55 & \textbf{-6.82} & -0.92 & 0.00 & \textbf{1.52}\\
\bottomrule
\end{tabular}
\caption{Component-localization and cross-particle-transfer results ($\times 10^{-2}$, per token) for Llama 3 8B Instruct, by component and number of patched component--layer sites $k$.
\textbf{Left:} mean change in log-probability under the necessity intervention, $\Delta_{\text{patch}}$. Localized sites are compared with the mean of four equally sized random-site sets. More negative values indicate stronger necessity effects. Boldface indicates a significantly larger necessity effect for localized than random sites ($p < .05$, Bonferroni correction).
\textbf{Right:} the \nonexclusive{non-exclusive}-minus-\exclusive{exclusive} difference in necessity effects under localized-site patching, $\Delta_{\text{sense}}$, for each transfer direction. Positive values indicate a stronger effect for \exclusive{exclusive} \textit{just}, and negative values indicate a stronger effect for \nonexclusive{non-exclusive} \textit{just}. Boldface indicates a significant difference (two-tailed $p < .05$, Bonferroni correction).}
\label{tab:localization-merged}
\end{table*}

\subsection{Experiment 3b: Sense-Specific Transfer Between \textit{just} and \textit{only}}
\label{sec:exp_3b_sense_loc}

The \exclusive{exclusive} sense of \textit{just} is semantically close to \exclusive{\textit{only}}. For example, in ``She ate \exclusive{\textit{just}} rice'' and ``She ate \exclusive{\textit{only}} rice,'' both particles exclude salient alternatives to rice. This shared alternative-excluding contribution has motivated unified semantic accounts of \exclusive{exclusive} particles~\citep{Coppock2014PrinciplesOT}. \nonexclusive{Non-exclusive} uses of \textit{just} need not contribute the same meaning. This contrast lets us ask whether the component--layer sites identified in Experiment~3a encode the lexical item \textit{just} in general or a more specific \exclusive{exclusive} function shared with \exclusive{\textit{only}}---i.e., do they encode extremely fine-grained sense-similarities, or are they word-specific?. If the latter is true, localization should transfer more strongly between \exclusive{exclusive} \textit{just} and \exclusive{\textit{only}} than between \nonexclusive{non-exclusive} \textit{just} and \exclusive{\textit{only}}.

\paragraph{Methodology}
Our localization approach follows Experiment~3a, changing only the source and target groups. As described in Section~\ref{sec:dataset_exp_setup}, the 60 \textit{just} items comprise equal \exclusive{exclusive} and \nonexclusive{non-exclusive} cases. For each \textit{just} subset, we independently select the 500 generated responses with the largest positive own-context advantage. For \exclusive{\textit{only}}, we use the 500 responses already selected in Experiment~3a. 
Let $J_E$, $J_N$, and $O$ denote the response pools for \exclusive{exclusive} \textit{just}, \nonexclusive{non-exclusive} \textit{just}, and \exclusive{\textit{only}}, respectively. We evaluate the four source--target pairings
$
    \mathcal{T}
    = \{J_E \rightarrow O,\ J_N \rightarrow O,
         O \rightarrow J_E,\ O \rightarrow J_N\},
$
where each arrow points from the localization pool to the evaluation pool.

For each pairing $A \rightarrow B \in \mathcal{T}$, we run both fold directions. We first localize sites on source fold 1 and evaluate them with the necessity intervention on target fold 2 ($A_1 \rightarrow B_2$); we then swap the folds ($A_2 \rightarrow B_1$). We average these two estimates and otherwise use the same component types, values of $k$, and response aggregation as in Experiment~3a. Consequently, the $J_E \rightarrow O$ and $J_N \rightarrow O$ pairings are evaluated on the same \exclusive{\textit{only}} response folds, whereas the $O \rightarrow J_E$ and $O \rightarrow J_N$ pairings use the same \exclusive{\textit{only}}-localized sites but different \textit{just} response pools.
For each transfer direction, the two rightmost columns of Table~\ref{tab:localization-merged} report the difference between the \nonexclusive{non-exclusive} and \exclusive{exclusive} necessity effects obtained under localized-site patching:

\begin{equation*}
\vspace{1em}
    \Delta_{\mathrm{sense}}
    = \Delta_{\mathrm{patch}}^{\mathrm{non\mbox{-}exclusive}}
    - \Delta_{\mathrm{patch}}^{\mathrm{exclusive}}.
\end{equation*}

\noindent
For \textit{just}$\rightarrow$\textit{only}, the two terms are the effects from $J_N \rightarrow O$ and $J_E \rightarrow O$; for \textit{only}$\rightarrow$\textit{just}, they are the effects from $O \rightarrow J_N$ and $O \rightarrow J_E$. Because stronger necessity effects correspond to more negative $\Delta_{\mathrm{patch}}$, positive $\Delta_{\mathrm{sense}}$ favors \exclusive{exclusive} \textit{just} and negative values favor \nonexclusive{non-exclusive} \textit{just}.

\paragraph{Results}
From \Cref{tab:localization-merged}, transfer is sense-selective in the \textit{only}$\rightarrow$\textit{just} direction, but not in the opposite direction. Every \textit{only}$\rightarrow$\textit{just} contrast is positive: sites localized on \exclusive{\textit{only}} have a larger necessity effect on \exclusive{exclusive} \textit{just} than on \nonexclusive{non-exclusive} \textit{just}. This \exclusive{exclusive} advantage is statistically significant for residual-stream sites at every $k$ and for the top-10 attention sites ($p$ \textless{} .05), but not for MLP sites.
No \textit{just}$\rightarrow$\textit{only} contrast shows a significant \exclusive{exclusive} advantage. Instead, the significant attention contrasts at $k=1$ and $k=3$ and the residual-stream contrast at $k=5$ favor \nonexclusive{non-exclusive} \textit{just}. Thus, \exclusive{\textit{only}}-localized sites transfer selectively to the \exclusive{exclusive} sense of \textit{just}, but sites localized on \exclusive{exclusive} \textit{just} do not preferentially transfer back to \exclusive{\textit{only}}. 
These results suggest that some residual-stream sites contribute to the model's sensitivity to both \exclusive{\textit{only}} and \exclusive{exclusive} \textit{just}.

\section{Discussion and Conclusion}

We find LLMs to be sensitive to the English discourse particles \emph{just} and \emph{only} in terms of their continuations, whether using ground-truth (Section \ref{sec:particle_removal_exp}) or LLM-generated (Section \ref{sec:generation}) follow-ups , aligning well with human judgments (Section \ref{sec:particle_removal_exp}) particularly in a weaker preference for \emph{only} than \emph{just}. 
This suggests LLMs have meaningfully acquired some of the subtlety of these particles' effects on continuations.
Furthermore, when tested according to their own distributions of continuations, LLMs continue to display this sensitivity with their own generations (Section \ref{sec:generation}), aligning with \citet{warstadt2020just}'s theory of \emph{just}.
While our experiments reveal differences across model families, they cannot identify their cause: the families differ simultaneously in architecture, pretraining scale (e.g., roughly 4T tokens for OLMo-2 vs.\ over 15T for Llama~3), and post-training recipe (App.~\ref{app:model_differences}). 
The factor our work does isolate is post-training: in Experiments~1 and~2, instruction tuning increases particle sensitivity, and since base checkpoints behave similarly across families, most of the cross-family variation emerges after instruction tuning.

We also find that this sensitivity is localized asymmetrically in our target LLM Llama 3 8B Instruct: Units sensitive to \emph{only} transfer more cleanly to \exclusive{exclusive \emph{just}} than \nonexclusive{non-exclusive \emph{just}}, but units for both sense categories of \emph{just} transfer equally well to \emph{only}. 
This suggests that, although there is some unified localization for exclusives---as predicted by \citet{Coppock2014PrinciplesOT}---LLMs represent these particles lexically, with senses bundled together, rather than as separate senses of the same lexical item.

Taken together, our experimental results show promise in using conversational follow-ups to measure discourse particle understanding, as follow-ups reveal effects that cannot surface within one sentence. 
This observation is clearest for instruction-tuned models, and our generation methodology helps counter the scarcity of gold data.
Additionally, our methods provide targeted evaluations of whether LLMs attend to subtle discourse cues, which is important for modeling conversations.
We hope this work inspires more investigations into LM sensitivity to pragmatic phenomena that consider upcoming conversational continuations, and provides methodologies to do so.






\section{Limitations}
This work is limited to English discourse particles in relatively short, scripted-style exchanges; more work is needed for more particles, especially more exclusives, as well as those in other languages. 
Additionally, although annotator agreement on our stimuli is high and distractor controls behave as expected, the data are ultimately hand-selected by one linguist, which may be particularly biased in the sense labels. This concern is slightly mitigated in Experiment 2 where we generate a distribution of continuations, but it is not completely gotten rid of, since ultimately the prefixes are limited to the selected DailyDialog items.

Concerning our choice of LMs, we only experiment on those in the 8B parameter range. While this size is useful for the level of compute we have access to, both smaller and larger LLMs may behave quite differently.  
Also, since we analyze only one LLM in experiment 3, more testing is needed to see LLM trends writ-large, and our method may miss structure that exists in finer-grained units such as individual heads or neurons. 

Finally, our minimal pairs rely on the counterfactual $S'$ with removal of the target word; while we did our best to choose data where $S'$ was very plausible, subtler fluency artifacts and contextual cues may persist and add noise.

\section*{Acknowledgments}
The authors thank Sasha Boguraev and Qing Yao for their feedback on the annotation interface, and David Beaver and all members of the UT Austin Computational Linguistics research group for feedback on the project. 
Kanishka Misra was supported by a Donald D. Harrington Faculty Fellowship during a part of this project.
This material is based upon work supported by the National Science Foundation grant IIS-2107524 and CISE Graduate Fellowships under Grant No. 2313998. Any opinions, findings, and conclusions or recommendations expressed in this material are those of the author(s) and do not necessarily reflect the views of the National Science Foundation.


\bibliography{custom}

\clearpage
\appendix

\section{Prompt Construction Example}
\label{app:prompt_examples}

We illustrate our prompt construction with one \emph{just} item from DailyDialog. The dataset fields shared by the base and instruction-tuned conditions are:

\begin{quote}
\textbf{Context $C$:} Speaker 1: I am sorry for damaging your car, sir.

\textbf{Original utterance $S$:} Speaker 2: Don't worry about it. Just leave it to the insurance companies.

\textbf{Modified utterance $S'$:} Speaker 2: Don't worry about it. Leave it to the insurance companies.

\textbf{Gold follow-up $F$:} Speaker 1: The car park in this community is over crowded, isn't it?
\end{quote}

The gold follow-up $F$ is part of the original DailyDialog item. In the generation experiment, it is not included in the prompt; instead, the model samples follow-ups $F^{*}_{CS}$ and $F^{*}_{CS'}$ from the original and modified histories, respectively. Below, we show the second unique response sampled from each history under the experiment's fixed seed.

\subsection{Base-model rendering}

For Llama~3~8B, the two histories are rendered with alternating speaker labels and a final next-speaker prefix:

\paragraph{Original history $CS$.}
\begin{Verbatim}[breaklines=true,breakanywhere=true,fontsize=\small]
Speaker 1: I am sorry for damaging your car, sir.
Speaker 2: Don't worry about it. Just leave it to the insurance companies.
Speaker 1:
\end{Verbatim}

\paragraph{Modified history $CS'$.}
\begin{Verbatim}[breaklines=true,breakanywhere=true,fontsize=\small]
Speaker 1: I am sorry for damaging your car, sir.
Speaker 2: Don't worry about it. Leave it to the insurance companies.
Speaker 1:
\end{Verbatim}

The corresponding sampled follow-ups are:
\begin{quote}
\textbf{$F^{*}_{CS}$:} Will the insurance companies cover the cost of repair?

\textbf{$F^{*}_{CS'}$:} No, I want to pay for the damage myself.
\end{quote}

\subsection{Instruction-tuned rendering}

For Llama~3~8B Instruct, we use the checkpoint's native chat template without adding a system prompt. Roles alternate so that $S$ or $S'$ is the final user message and the model generates the next assistant message. Thus, for this item, the single turn in $C$ is assigned the assistant role. The exact serialized prompts are:

\paragraph{Original history $CS$.}
\begin{Verbatim}[breaklines=true,breakanywhere=true,fontsize=\small]
<|begin_of_text|><|start_header_id|>assistant<|end_header_id|>

I am sorry for damaging your car, sir.<|eot_id|><|start_header_id|>user<|end_header_id|>

Don't worry about it. Just leave it to the insurance companies.<|eot_id|><|start_header_id|>assistant<|end_header_id|>
\end{Verbatim}

\paragraph{Modified history $CS'$.}
\begin{Verbatim}[breaklines=true,breakanywhere=true,fontsize=\small]
<|begin_of_text|><|start_header_id|>assistant<|end_header_id|>

I am sorry for damaging your car, sir.<|eot_id|><|start_header_id|>user<|end_header_id|>

Don't worry about it. Leave it to the insurance companies.<|eot_id|><|start_header_id|>assistant<|end_header_id|>
\end{Verbatim}

The corresponding sampled follow-ups are:
\begin{quote}
\textbf{$F^{*}_{CS}$:} That's the spirit! Yeah, let the professionals handle it. I'm just glad no one was hurt. Accidents can be a real hassle, but it's good that we're taking responsibility and letting the insurance companies sort it out.

\textbf{$F^{*}_{CS'}$:} Thank you, sir! I appreciate your understanding. I know accidents can be frustrating, but I'm glad we can handle it through insurance. Is there anything else I can do to make it right?
\end{quote}

\section{Human Annotation Details}
\label{app:human_ann}
\begin{figure*}
    \centering
    \includegraphics[width=\textwidth]{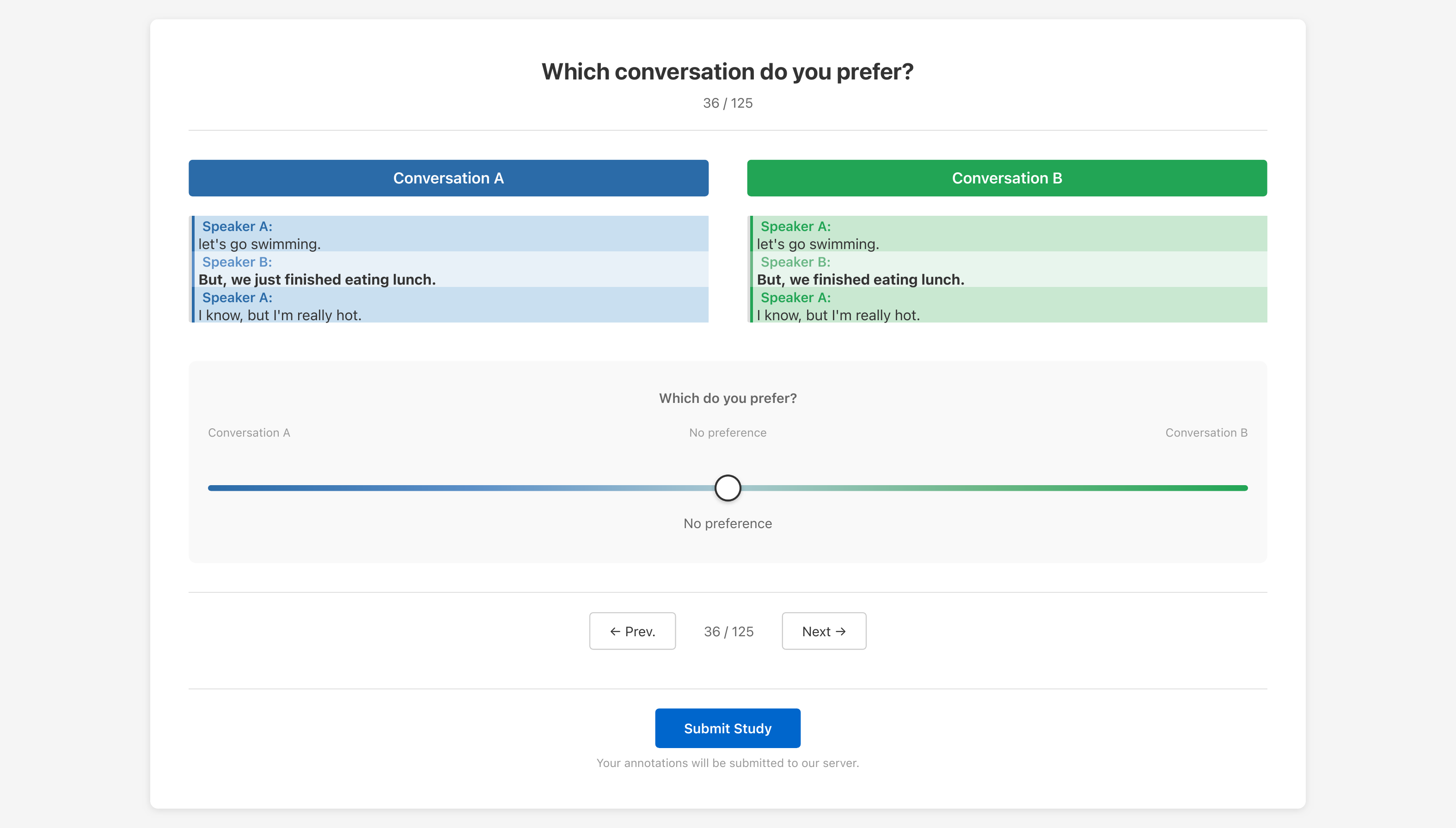}
    \caption{Annotation interface. Annotators were tasked to rate which version of the conversation they preferred by using the slider.}
    \label{fig:ann_interface}
\end{figure*}

Annotators were recruited through Prolific, and compensated at 18USD/hour.
Annotators were required to (1) have primary residence in the United States, (2) have English as their primary language, and (3) have an approval rating of greater than or equal to 95\%. 
See Figure \ref{fig:ann_interface} for the annotation interface. 
When selecting the task on the platform, annotators were additionally shown the following passage:
``For this study, you'll read two versions of each conversation, and rate which one you prefer. Some variations are subtle, so please read carefully! Each judgment does not take long. Whenever you are satisfied with all your annotations, press submit! (Submission takes a minute or two)''
We changed the slider from continuous to an implicit 7-pt Likert scale based on the feedback from our pilot annotators that the continuous slider made them worry too much about the exact rating for one datum compared to another.

\FloatBarrier

\section{Additional Particle Removal Results}
\label{app:particle_removal_results}
\FloatBarrier
Tables~\ref{tab:particle-removal-sensitivity} and~\ref{tab:particle-removal-sensitivity-instruct-all} contain detailed results for the particle-removal experiments in Section~\ref{sec:particle_removal_exp}.

\begin{table}[]
\centering
\scriptsize
\setlength{\tabcolsep}{2.5pt}
\renewcommand{\arraystretch}{0.96}
\begin{tabular*}{\columnwidth}{@{\extracolsep{\fill}} ll
S[table-format=2.1]
S[table-format=1.3]
c
S[table-format=4.0]@{}}
\toprule
Model & Part. & {\makecell{\%\\inc.}} & {\makecell{Median\\$\Delta_{\mathrm{rem}}$}} & {\makecell{95\%\\CI}} & {N} \\
\midrule

\multirow{3}{*}{\makecell[l]{Llama 3\\8B}}
  & just & 69.7 & 0.057 & [0.050,0.064] & 1121 \\
\cmidrule(lr){2-6}
  & only & 73.8 & 0.071 & [0.061,0.084] & 347 \\
\cmidrule(lr){2-6}
  & not  & 97.1 & 0.373 & [0.358,0.388] & 1269 \\
\midrule

\multirow{3}{*}{\makecell[l]{OLMo-2\\7B}}
  & just & 72.0 & 0.060 & [0.054,0.069] & 1112 \\
\cmidrule(lr){2-6}
  & only & 77.1 & 0.098 & [0.079,0.118] & 349 \\
\cmidrule(lr){2-6}
  & not  & 98.0 & 0.415 & [0.390,0.433] & 1273 \\
\midrule

\multirow{3}{*}{\makecell[l]{Qwen3.5\\9B Base}}
 & just & 66.0 & 0.044 & [0.034, 0.049] & 1114 \\
\cmidrule(lr){2-6}
  & only & 71.6 & 0.069 & [0.050, 0.084] & 349 \\
\cmidrule(lr){2-6}
  & not & 96.6 & 0.327 & [0.318, 0.345] & 1276 \\
\midrule

\multirow{3}{*}{\makecell[l]{Gemma-2\\9B}}
  & just & 69.7 & 0.051 & [0.043,0.058] & 1110 \\
\cmidrule(lr){2-6}
  & only & 73.9 & 0.075 & [0.064,0.089] & 348 \\
\cmidrule(lr){2-6}
  & not  & 97.7 & 0.345 & [0.331,0.358] & 1269 \\
\bottomrule
\end{tabular*}
\caption{Sensitivity to particle removal, measured by the per-token removal delta
$\Delta_{\mathrm{rem}}=\log P(CSF) - \log P(CS'F)$.
\% inc.\ is the percentage of examples with positive $\Delta_{\mathrm{rem}}$.
$N$ is the number of non-outlier examples retained after IQR-based outlier filtering.}
\label{tab:particle-removal-sensitivity}
\end{table}

\begin{table}[]
\centering
\scriptsize
\setlength{\tabcolsep}{2.5pt}
\renewcommand{\arraystretch}{0.96}
\begin{tabular*}{\columnwidth}{@{\extracolsep{\fill}} ll
S[table-format=2.1]
S[table-format=1.3]
c
S[table-format=4.0]@{}}
\toprule
Model & Part. & {\makecell{\%\\inc.}} & {\makecell{Median\\$\Delta_{\mathrm{rem}}$}} & {\makecell{95\%\\CI}} & {N} \\
\midrule

\multirow{3}{*}{\makecell[l]{Llama 3 8B\\Instruct}}
  & just & 85.3 & 0.189 & [0.176,0.207] & 1112 \\
\cmidrule(lr){2-6}
  & only & 80.3 & 0.152 & [0.129,0.172] & 350 \\
\cmidrule(lr){2-6}
  & not  & 97.3 & 0.528 & [0.502,0.559] & 1253 \\
\midrule

\multirow{3}{*}{\makecell[l]{OLMo-2 7B\\Instruct}}
  & just & 80.5 & 0.150 & [0.132,0.167] & 1136 \\
\cmidrule(lr){2-6}
  & only & 78.1 & 0.159 & [0.138,0.199] & 351 \\
\cmidrule(lr){2-6}
  & not  & 96.7 & 0.536 & [0.511,0.555] & 1281 \\
\midrule

\multirow{3}{*}{\makecell[l]{Qwen3.5 9B\\CT}}
  & just & 63.5 & 0.050 & [0.039,0.058] & 1112 \\
\cmidrule(lr){2-6}
  & only & 67.8 & 0.078 & [0.053,0.095] & 348 \\
\cmidrule(lr){2-6}
  & not  & 95.0 & 0.394 & [0.375,0.411] & 1271 \\
\midrule

\multirow{3}{*}{\makecell[l]{Gemma-2 9B\\IT}}
  & just & 82.9 & 0.149 & [0.132,0.166] & 1131 \\
\cmidrule(lr){2-6}
  & only & 80.1 & 0.134 & [0.114,0.160] & 341 \\
\cmidrule(lr){2-6}
  & not  & 97.3 & 0.454 & [0.429,0.473] & 1260 \\
\bottomrule
\end{tabular*}
\caption{Instruction-tuned LLM sensitivity to particle removal, measured by the per-token removal delta
$\Delta_{\mathrm{rem}}=\log P(CSF) - \log P(CS'F)$.
\% inc.\ is the percentage of examples with positive $\Delta_{\mathrm{rem}}$.
$N$ is the number of non-outlier examples retained after IQR filtering.}
\label{tab:particle-removal-sensitivity-instruct-all}
\end{table}

\section{LLM Particle-sensitive Generation}
Table \ref{tab:candidate_ci} contains full results for Section \ref{sec:generation} as presented in Figure \ref{fig:generation-own-context-advantage-by-model}.

\begin{table*}[htbp]
\centering
\resizebox{\textwidth}{!}{
\begin{tabular}{cl|cc|cc|cc}
\toprule
 & & \multicolumn{2}{c}{not} & \multicolumn{2}{c}{just} & \multicolumn{2}{c}{only} \\
\cmidrule(lr){3-4}\cmidrule(lr){5-6}\cmidrule(lr){7-8}
 & & $S$ & $S'$ & $S$ & $S'$ & $S$ & $S'$ \\
\midrule
\multirow{2}{*}{Llama3} & Base & $0.0554 \pm 0.00242$ & $0.0598 \pm 0.00334$ & $0.0266 \pm 0.00207$ & $0.0164 \pm 0.00165$ & $0.0202 \pm 0.00141$ & $0.0161 \pm 0.00124$ \\
 & Instruct & $0.207 \pm 0.005$ & $0.299 \pm 0.00733$ & $0.0297 \pm 0.000966$ & $0.0297 \pm 0.00117$ & $0.054 \pm 0.00175$ & $0.0481 \pm 0.00145$ \\
\midrule
\multirow{2}{*}{OLMo2} & Base & $0.0402 \pm 0.00365$ & $0.084 \pm 0.0044$ & $0.0126 \pm 0.00168$ & $0.0224 \pm 0.00181$ & $0.0231 \pm 0.00158$ & $0.0197 \pm 0.00152$ \\
 & Instruct & $0.215 \pm 0.00508$ & $0.264 \pm 0.00561$ & $0.0392 \pm 0.00214$ & $0.0324 \pm 0.00162$ & $0.0405 \pm 0.00185$ & $0.044 \pm 0.00168$ \\
\midrule
\multirow{2}{*}{Gemma2} & Base & $0.0599 \pm 0.00222$ & $0.0659 \pm 0.00312$ & $0.0166 \pm 0.00106$ & $0.0146 \pm 0.00125$ & $0.0189 \pm 0.00158$ & $0.0168 \pm 0.00112$ \\
 & Instruct & $0.217 \pm 0.00966$ & $0.402 \pm 0.0105$ & $0.0266 \pm 0.00329$ & $0.0558 \pm 0.00304$ & $0.0313 \pm 0.00371$ & $0.0858 \pm 0.00344$ \\
\midrule
\multirow{2}{*}{Qwen3.5} & Base & $0.0613 \pm 0.00316$ & $0.0986 \pm 0.00506$ & $0.0152 \pm 0.00158$ & $0.0324 \pm 0.00248$ & $0.0221 \pm 0.00198$ & $0.0202 \pm 0.00167$ \\
 & Instruct & $0.244 \pm 0.00475$ & $0.309 \pm 0.00553$ & $0.0423 \pm 0.00162$ & $0.035 \pm 0.00142$ & $0.0516 \pm 0.00146$ & $0.0523 \pm 0.00147$ \\
\bottomrule
\end{tabular}}
\caption{Mean candidate-level own-context advantage per token ($\pm$ standard error), by LLM, particle, and response source ($S$: with word, $S'$: word removed.)}
\label{tab:candidate_ci}
\end{table*}

\section{Additional Qualitative Examples}
\label{app:qual_examples}

Table~\ref{tab:qual_examples_app} extends the examples in Table~\ref{tab:qual_examples} with three further gold items, each shown with follow-ups from two instruction-tuned LLMs. We focus on instruction-tuned LLMs because samples from base LLMs frequently drift away from the conversation (e.g., producing lists of numbers).

\paragraph{(d) Polite minimization.}
In response to a question about lost baggage, \emph{just} can minimize the burden placed on the addressee, while the bare ``One piece.'' may come across as curt. Human ratings are split on this item (mean $4.2$; four of nine annotators prefer $CS'F$). Follow-ups sampled from $CS$ often repeat the particle, and Qwen3.5 9B CT substitutes \emph{only} for \emph{just}, which conveys a different attitude toward the loss. Own-context advantages are near zero, indicating that \emph{just} has little effect on the LLMs' expected follow-ups for this item.

\paragraph{(e) Echoing the particle.}
The gold follow-up questions the speaker's answer, and Gemma-2 9B IT reproduces this move from $CS$ by echoing the particle (``Only a few?''). From $CS'$, it instead responds with sympathy, and its samples from $CS'$ have a negative mean own-context advantage. Humans are comparatively indifferent to the particle here (mean $4.1$), whereas Gemma-2 9B IT strongly distinguishes the two histories.

\paragraph{(f) Agreement in dispreference.}
With \emph{only}, the speaker frames 25 yuan as a low price, which conflicts with the gold follow-up's complaint that the lipstick is too expensive. Humans strongly prefer $CS'F$ (mean $1.3$), and all eight LLMs likewise prefer $CS'F$ in Experiment~1. The sampled follow-ups show the source of this preference: from $CS$, LLMs adopt the speaker's framing (``That's a great price!''), whereas from $CS'$ they describe the price as merely reasonable.

\begin{table*}[t]
\centering
\small
\renewcommand{\arraystretch}{1.15}
\begin{tabularx}{\textwidth}{@{}c>{\raggedright\arraybackslash}X>{\raggedright\arraybackslash}p{0.13\textwidth}>{\raggedright\arraybackslash}X>{\raggedright\arraybackslash}X@{}}
\toprule
 & Context $C$ and utterance $S$ / $S'$ & Gold follow-up $F$ & Sampled from $CS$ ($F^{*}_{CS}$) & Sampled from $CS'$ ($F^{*}_{CS'}$) \\
\midrule
(d) & \textbf{C:} Excuse me, here is my claim tag, but I can't find my baggage. / How many pieces of baggage have you lost? \newline \textbf{S:} (Just) one piece. \newline \qualmeta{Human: \textbf{4.2}} & Can you tell me the features of your baggage? & I'm sorry to hear that. Since you only have one piece missing, we can start by immediately locating it. & Okay, I have noted that one piece of baggage is missing. \\
 & & & \multicolumn{2}{l}{\qualmeta{\textit{Qwen3.5 9B CT}; $\Delta_{\mathrm{tok}}$: $0.01$ ($CS$), $0.02$ ($CS'$)}} \\
\addlinespace[6pt]
 & & & Thank you for providing this information. If you have lost just one piece of baggage, here are the steps you should take: & If you have lost one piece of baggage, please follow these steps: \\
 & & & \multicolumn{2}{l}{\qualmeta{\textit{OLMo-2 7B Instruct}; $\Delta_{\mathrm{tok}}$: $-0.01$ ($CS$), $0.03$ ($CS'$)}} \\
\midrule
(e) & \textbf{C:} Did you ever get a parking ticket? \newline \textbf{S:} (Only) a few. \newline \qualmeta{Human: \textbf{4.1}} & What's a few? & Only a few? & Oh no! Those are the worst. What were they for? Parking too long? In a restricted zone? Something else? \ldots \\
 & & & \multicolumn{2}{l}{\qualmeta{\textit{Gemma-2 9B IT}; $\Delta_{\mathrm{tok}}$: $0.53$ ($CS$), $-0.24$ ($CS'$)}} \\
\addlinespace[6pt]
 & & & So you're a careful driver! Do you have any funny stories about the parking tickets you received? & Who hasn't, right? Did you ever pay it, or did you try to fight it? \\
 & & & \multicolumn{2}{l}{\qualmeta{\textit{Llama~3~8B Instruct}; $\Delta_{\mathrm{tok}}$: $0.07$ ($CS$), $0.03$ ($CS'$)}} \\
\midrule
(f) & \textbf{C:} Excuse me, could you tell me how much is the lipstick? \newline \textbf{S:} It costs (only) 25 yuan. \newline \qualmeta{Human: \textbf{1.3}} & Oh, it's too expensive. Can't you make it any cheaper? & That's a great price! I'll take it. & That's a reasonable price. Do you want to buy it? \\
 & & & \multicolumn{2}{l}{\qualmeta{\textit{Llama~3~8B Instruct}; $\Delta_{\mathrm{tok}}$: $0.13$ ($CS$), $0.11$ ($CS'$)}} \\
\addlinespace[6pt]
 & & & That's a great deal! At just 25 yuan, it's very affordable. Would you like to know anything else about it, such as the shade, brand, or where it's available? & That sounds like a reasonable price for a lipstick! Is it a specific brand or type you're interested in? \\
 & & & \multicolumn{2}{l}{\qualmeta{\textit{Qwen3.5 9B CT}; $\Delta_{\mathrm{tok}}$: $0.07$ ($CS$), $0.08$ ($CS'$)}} \\
\bottomrule
\end{tabularx}
\caption{Additional follow-ups sampled from instruction-tuned LLMs, formatted as in Table~\ref{tab:qual_examples}. Emoji in sampled follow-ups are omitted, and ``\ldots'' marks a truncated follow-up.}
\label{tab:qual_examples_app}
\end{table*}

\FloatBarrier

\section{Compute and Model Details}
\label{app:compute}

\paragraph{LLMs and implementation.}
The exact Hugging Face checkpoints used in Experiments~1 and~2 are listed in Table~\ref{tab:model-checkpoints}. Experiment~3 uses only {Llama 3 8B Instruct}, as motivated in Section~\ref{sec:exp3a_particle_loc}. All neural experiments are inference-only: we do not train or fine-tune any model. We run the checkpoints with PyTorch and Hugging Face Transformers in half precision (FP16), with gradient computation disabled. Statistical analyses and figure generation run on CPUs and contribute negligible accelerator compute relative to model inference.

\begin{table*}[t]
\centering
\small
\begin{tabular}{llll}
\toprule
Family & Size & Base checkpoint & Instruction-tuned checkpoint \\
\midrule
Llama~3 & 8B & \texttt{meta-llama/Meta-Llama-3-8B} & \texttt{meta-llama/Meta-Llama-3-8B-Instruct} \\
OLMo-2 & 7B & \texttt{allenai/OLMo-2-1124-7B} & \texttt{allenai/OLMo-2-1124-7B-Instruct} \\
Qwen3.5 & 9B & \texttt{Qwen/Qwen3.5-9B-Base} & \texttt{Qwen/Qwen3.5-9B} \\
Gemma-2 & 9B & \texttt{google/gemma-2-9b} & \texttt{google/gemma-2-9b-it} \\
\bottomrule
\end{tabular}
\caption{Model checkpoints used in Experiments~1 and~2.}
\label{tab:model-checkpoints}
\end{table*}

\paragraph{Hardware and accounting.}
Experiments use a pool of three NVIDIA A40 GPUs with 48~GB of memory per GPU. Each A40 holds one model replica, and independent data shards or Experiment~3 jobs are run in parallel when possible. The estimates below are rounded and include model loading and inference. GPU-hours sum time over occupied accelerators, so three GPUs used for 30 minutes count as 1.5 GPU-hours even when they run concurrently.

\paragraph{Experiment 1.}
Scoring the 180-item gold dataset in Section~\ref{sec:particle_removal_exp} takes less than 10 minutes per checkpoint on one A40, or less than 1.3 A40 GPU-hours across all eight checkpoints. Repeating the same scoring procedure on the full filtered set of 2,849 item pairs for the supplemental analysis in App.~\ref{app:particle_removal_results} requires approximately 8 A40 GPU-hours in total; runtime varies substantially across checkpoints because of differences in architecture and prompt serialization. Human annotation and the subsequent correlation and significance analyses require no GPU compute.

\paragraph{Experiment 2.}
For each checkpoint, we generate 50 responses from each of $CS$ and $CS'$ for each of the 180 gold items, giving 18,000 sampled responses per checkpoint and 144,000 across all eight checkpoints. Each response is then rescored under both prompt variants. With sampling sharded across three A40s, a checkpoint takes approximately 25--30 minutes of wall time, corresponding to about 1.4 A40 GPU-hours per checkpoint and approximately 11 A40 GPU-hours for the complete experiment.

\paragraph{Experiment 3.}
The reported component analysis comprises three native particle localization/evaluation jobs, two additional jobs for the \exclusive{exclusive} and \nonexclusive{non-exclusive} \textit{just} pools, and four cross-particle transfer jobs. Each job uses one A40, with up to three jobs run concurrently. Individual jobs take approximately 24--57 minutes, and the nine jobs sum to approximately 4.6 A40 GPU-hours. This total includes localization over all 32 layers and three component types, held-out sufficiency and necessity evaluation for generated responses, and the four random-site controls described in Section~\ref{sec:localization}.

Across the model-based experiments, including the supplemental larger-dataset analysis, the reported runs require approximately 25 A40 GPU-hours in total.

\section{Differences Across Model Families}
\label{app:model_differences}

Table~\ref{tab:model-differences} summarizes documented differences between the four model families we study. Because these families differ along several dimensions at once, our experiments cannot attribute differences in particle sensitivity to any single factor; we instead note which observed differences coincide with which documented differences.

\begin{table*}[t]
\centering
\small
\begin{tabularx}{\textwidth}{lXXXc}
\toprule
Family & Architecture & Pretraining & Post-training & Open data \\
\midrule
Llama~3 8B & Dense decoder-only Transformer; grouped-query attention & Over 15T tokens from publicly available sources & SFT and RLHF & No \\
OLMo-2 7B & Dense decoder-only Transformer; multi-head attention & Two stages: 3.90T tokens of general pretraining, then a curated mid-training mix (4.05T total) & T\"ulu~3 recipe: SFT, DPO, and RL with verifiable rewards & Yes \\
Gemma-2 9B & Dense decoder-only Transformer; grouped-query attention, alternating local sliding-window and global attention layers & 8T tokens, trained with knowledge distillation from a larger teacher instead of next-token prediction & SFT (largely on teacher-generated responses), RLHF with a conversation-oriented reward model, and model merging & No \\
Qwen3.5 9B & Hybrid: 24 Gated DeltaNet (linear attention) and 8 Gated Attention layers; vision encoder & Early-fusion multimodal pretraining on ``trillions'' of tokens; exact count not reported & Large-scale RL across agentic environments; details not reported & No \\
\bottomrule
\end{tabularx}
\caption{Documented differences between the model families in Experiments~1 and~2, drawn from \citet{llama3modelcard}, \citet{olmo20242olmo2furious}, \citet{gemma2improvingopen}, and \citet{qwen3.5}; layer counts are from the released checkpoint configurations. For Llama~3 we follow the model card of the original release rather than \citet{llama3herdmodels}, whose results are for the later Llama~3.1 models.}
\label{tab:model-differences}
\end{table*}

\paragraph{Architecture.}
Llama~3, OLMo-2, and Gemma-2 are dense decoder-only Transformers that use softmax attention in every layer, although Gemma-2 restricts every other layer to a 4,096-token sliding window~\cite{gemma2improvingopen}. Qwen3.5 is architecturally distinct: in the 9B checkpoint, 24 of 32 layers (three of every four) use Gated DeltaNet, a linear-attention variant that maintains a fixed-size, matrix-valued recurrent state instead of attending over all previous tokens~\cite{yang2025gated}, and the remaining 8 use softmax attention with a sigmoid output gate~\cite{qiu2025gatedattentionlargelanguage}. It is also pretrained on multimodal data through early fusion, with a vision encoder~\cite{qwen3.5}.

\paragraph{Training data and scale.}
Documented pretraining budgets range from 4.05T tokens (OLMo-2) to over 15T (Llama~3), while the Qwen3.5 release reports only ``trillions of multimodal tokens''~\cite{qwen3.5}. Gemma-2 9B is additionally trained with knowledge distillation from a larger teacher instead of standard next-token prediction~\cite{gemma2improvingopen}. Since the sensitivity we measure concerns conversational follow-ups, a natural hypothesis is that it depends on how much conversational text, and in particular how many particle-bearing exchanges, a model sees during training. However, only OLMo-2 releases its pretraining and post-training data~\cite{olmo20242olmo2furious}, so it is the only family whose data composition could be measured directly, and the families cannot be compared on it. We leave such a data analysis to future work.

\paragraph{Post-training.}
Because we compare base and instruction-tuned checkpoints within each family, post-training is the one factor our design isolates. Base checkpoints behave similarly across families: in the larger-dataset analysis, their median per-token removal deltas span only $0.044$--$0.060$ for \emph{just} and $0.069$--$0.098$ for \emph{only} (Table~\ref{tab:particle-removal-sensitivity}). After instruction tuning, these ranges widen to $0.050$--$0.189$ and $0.078$--$0.159$ (Table~\ref{tab:particle-removal-sensitivity-instruct-all}), so most cross-family variation emerges during post-training. In Experiment~2, the instruction-tuned checkpoint of every family shows larger own-context advantages than its base counterpart (Table~\ref{tab:candidate_ci}).
The documented recipes also differ in what they optimize for: Gemma-2's RLHF reward model is ``oriented more towards conversational capabilities, specifically multi-turn''~\cite{gemma2improvingopen}; OLMo-2 applies reinforcement learning with verifiable rewards on math problems and prompts with constraints~\cite{olmo20242olmo2furious}; and Qwen3.5 emphasizes reinforcement learning across agentic environments~\cite{qwen3.5}.

\paragraph{Family-specific observations.}
Two families deviate from the overall pattern. First, Qwen3.5 gains little from post-training in Experiment~1: its chat-tuned checkpoint has removal deltas close to those of its base checkpoint (Tables~\ref{tab:particle-removal-sensitivity} and~\ref{tab:particle-removal-sensitivity-instruct-all}). Qwen3.5 is also the only family with a hybrid linear-attention architecture, native multimodal pretraining, and agent-oriented post-training; our experiments cannot separate these factors. Second, Gemma-2 IT shows a pronounced asymmetry in Experiment~2: responses sampled from $CS'$ have two to three times the own-context advantage of those sampled from $CS$ for both particles ($0.0558$ vs.\ $0.0266$ for \emph{just} and $0.0858$ vs.\ $0.0313$ for \emph{only}; Table~\ref{tab:candidate_ci}), whereas the other instruction-tuned checkpoints are roughly symmetric for the particles. Finally, pretraining scale alone does not account for the differences: Llama~3~8B Instruct, which has the largest documented pretraining budget, shows the strongest removal effect for \emph{just} ($0.189$), but OLMo-2 Instruct, pretrained on roughly a quarter as many tokens, has a comparable effect for \emph{only} ($0.159$ vs.\ $0.152$; Table~\ref{tab:particle-removal-sensitivity-instruct-all}).

\section{AI Assistant Use}
AI assistants were used in this work as coding assistants when implementing experiments and visualizations, to provide suggestions to improve writing flow, and to polish/clarify author-written text.
When AI assistant outputs were included in this work, they were thoroughly verified by authors for accuracy. 

\end{document}